\documentclass{article}

\usepackage{times}
\usepackage{graphicx} % more modern
\usepackage{subfigure}

\usepackage{natbib}

\usepackage{algorithm}
\usepackage{algorithmic}
\usepackage[algo2e,linesnumbered,lined,boxed,vlined]{algorithm2e}

\usepackage{amsfonts}
\usepackage{amssymb}

\usepackage{multirow}
\usepackage{caption}

\newcommand{\indep}{\perp\!\!\!\perp}

\usepackage{hyperref}

\usepackage[accepted]{icml2016}

\usepackage{comment}
\usepackage{amsthm}
\usepackage{mathtools}
\usepackage{tabularx}
\usepackage{multirow}

\def\b{\ensuremath\boldsymbol}

\usepackage{lastpage}
\usepackage{fancyhdr}
\usepackage{url}
\icmltitlerunning{Linear and Quadratic Discriminant Analysis: Tutorial}

\begin{document}

\twocolumn[
\icmltitle{Linear and Quadratic Discriminant Analysis: Tutorial}

\icmlauthor{Benyamin Ghojogh}{bghojogh@uwaterloo.ca}
\icmladdress{Department of Electrical and Computer Engineering, 
\\Machine Learning Laboratory, University of Waterloo, Waterloo, ON, Canada}
% \icmlauthor{Fakhri Karray}{karray@uwaterloo.ca}
% \icmladdress{Department of Electrical and Computer Engineering, 
% \\Centre for Pattern Analysis and Machine Intelligence, University of Waterloo, Waterloo, ON, Canada}
\icmlauthor{Mark Crowley}{mcrowley@uwaterloo.ca}
\icmladdress{Department of Electrical and Computer Engineering, 
\\Machine Learning Laboratory, University of Waterloo, Waterloo, ON, Canada}

\icmlkeywords{Tutorial, Locally Linear Embedding}

\vskip 0.3in
]

\begin{abstract}
This tutorial explains Linear Discriminant Analysis (LDA) and Quadratic Discriminant Analysis (QDA) as two fundamental classification methods in statistical and probabilistic learning. We start with the optimization of decision boundary on which the posteriors are equal. Then, LDA and QDA are derived for binary and multiple classes. The estimation of parameters in LDA and QDA are also covered. Then, we explain how LDA and QDA are related to metric learning, kernel principal component analysis, Mahalanobis distance, logistic regression, Bayes optimal classifier, Gaussian naive Bayes, and likelihood ratio test. We also prove that LDA and Fisher discriminant analysis are equivalent. We finally clarify some of the theoretical concepts with simulations we provide.
\end{abstract}

\section{Introduction}

Assume we have a dataset of \textit{instances} $\{(\b{x}_i, y_i)\}_{i=1}^n$ with sample size $n$ and dimensionality $\b{x}_i  \in \mathbb{R}^d$ and $y_i \in \mathbb{R}$. The $y_i$'s are the class labels. We would like to \textit{classify} the space of data using these instances.
Linear Discriminant Analysis (LDA) and Quadratic discriminant Analysis (QDA) \cite{friedman2001elements} are two well-known \textit{supervised classification} methods in statistical and probabilistic learning.
This paper is a tutorial for these two classifiers where the theory for binary and multi-class classification are detailed. Then, relations of LDA and QDA to metric learning, kernel Principal Component Analysis (PCA), Fisher Discriminant Analysis (FDA), logistic regression, Bayes optimal classifier, Gaussian naive Bayes, and Likelihood Ratio Test (LRT) are explained for better understanding of these two fundamental methods.
Finally, some experiments on synthetic datasets are reported and analyzed for illustration.

\section{Optimization for the Boundary of Classes}

First suppose the data is one dimensional, $x \in \mathbb{R}$.
Assume we have two classes with the Cumulative Distribution Functions (CDF) $F_1(x)$ and $F_2(x)$, respectively. Let the Probability Density Functions (PDF) of these CDFs be:
\begin{align}
& f_1(x) = \frac{\partial F_1(x)}{\partial x}, \\
& f_2(x) = \frac{\partial F_2(x)}{\partial x}, 
\end{align}
respectively. 

We assume that the two classes have normal (Gaussian) distribution which is the most common and default distribution in the real-world applications. 
The mean of one of the two classes is greater than the other one; we assume $\mu_1 < \mu_2$. An instance $x \in \mathbb{R}$ belongs to one of these two classes:
\begin{align}
x \sim 
\left\{
\begin{array}{ll}
  \mathcal{N}(\mu_1, \sigma_1^2), & \text{if } x \in \mathcal{C}_1, \\
  \mathcal{N}(\mu_2, \sigma_2^2), & \text{if } x \in \mathcal{C}_2,
\end{array}
\right.
\end{align}
where $\mathcal{C}_1$ and $\mathcal{C}_2$ denote the first and second class, respectively.

\begin{figure}[!t]
\centering
\includegraphics[width=2.8in]{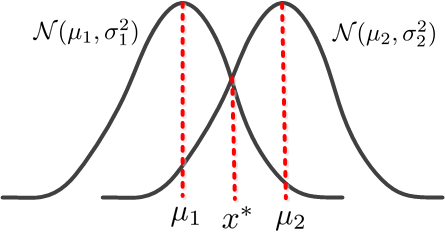}
\caption{Two Gaussian density functions where they are equal at the point $x^*$.}
\label{figure_two_Gaussians}
\end{figure}

For an instance $x$, we may have an error in estimation of the class it belongs to. At a point, which we denote by $x^*$, the probability of the two classes are equal; therefore, the point $x^*$ is on the boundary of the two classes. As we have $\mu_1 < \mu_2$, we can say $\mu_1 < x^* < \mu_2$ as shown in Fig. \ref{figure_two_Gaussians}.
Therefore, if $x < x^*$ or $x > x^*$ the instance $x$ belongs to the first and second class, respectively. Hence, estimating $x < x^*$ or $x > x^*$ for belonging to the second and first class, respectively, is an error in estimation of the class.
This probability of the error can be stated as:
\begin{align}
\mathbb{P}(\text{error}) = \mathbb{P}(x > x^* , x \in \mathcal{C}_1) + \mathbb{P}(x < x^* , x \in \mathcal{C}_2).
\end{align}
As we have $\mathbb{P}(A,B) = \mathbb{P}(A|B)\, \mathbb{P}(B)$, we can say:
\begin{equation}\label{equation_probability_error}
\begin{aligned}
\mathbb{P}(\text{error}) =\, &\mathbb{P}(x > x^* \,|\, x \in \mathcal{C}_1)\, \mathbb{P}(x \in \mathcal{C}_1) \\
&+ \mathbb{P}(x < x^* \,|\, x \in \mathcal{C}_2)\, \mathbb{P}(x \in \mathcal{C}_2),
\end{aligned}
\end{equation}
which we want to minimize:
\begin{align}\label{equation_optimization_probability_error}
\underset{x^*}{\text{minimize}} ~~~ \mathbb{P}(\text{error}),
\end{align}
by finding the best boundary of classes, i.e., $x^*$.

According to the definition of CDF, we have:
\begin{align}
& \mathbb{P}(x < c , x \in \mathcal{C}_1) = F_1(c), \nonumber \\
& \implies \mathbb{P}(x > x^* , x \in \mathcal{C}_1) = 1 - F_1(x^*), \\
& \mathbb{P}(x < x^* , x \in \mathcal{C}_2) = F_2(x^*).
\end{align}
According to the definition of PDF, we have:
\begin{align}
& \mathbb{P}(x \in \mathcal{C}_1) = f_1(x) = \pi_1, \\
& \mathbb{P}(x \in \mathcal{C}_2) = f_2(x) = \pi_2,
\end{align}
where we denote the priors $f_1(x)$ and $f_2(x)$ by $\pi_1$ and $\pi_2$, respectively. 

Hence, Eqs. (\ref{equation_probability_error}) and (\ref{equation_optimization_probability_error}) become:
\begin{align}
\underset{x^*}{\text{minimize}} ~~~ \big(1 - F_1(x^*)\big)\, \pi_1 + F_2(x^*)\, \pi_2.
\end{align}
We take derivative for the sake of minimization:
\begin{align}
&\frac{\partial\, \mathbb{P}(\text{error})}{\partial x^*} = -f_1(x^*)\, \pi_1 + f_2(x^*)\, \pi_2 \overset{\text{set}}{=} 0, \nonumber \\
&\implies f_1(x^*)\, \pi_1 = f_2(x^*)\, \pi_2. \label{equation_boundary_first_equation_1}
\end{align}
Another way to obtain this expression is equating the posterior probabilities to have the equation of the boundary of classes:
\begin{align}\label{equation_posteriors_equal}
\mathbb{P}(x \in \mathcal{C}_1 \,|\, X=x) \overset{\text{set}}{=} \mathbb{P}(x \in \mathcal{C}_2 \,|\, X=x).
\end{align}
According to Bayes rule, the \textit{posterior} is:
\begin{align}
\mathbb{P}(x \in \mathcal{C}_1 \,|\, X=x) &= \frac{\mathbb{P}(X=x \,|\, x \in \mathcal{C}_1)\, \mathbb{P}(x \in \mathcal{C}_1)}{\mathbb{P}(X=x)} \nonumber \\
&= \frac{f_1(x)\, \pi_1}{\sum_{k=1}^{|\mathcal{C}|} \mathbb{P}(X=x \,|\, x \in \mathcal{C}_k)\, \pi_k}, \label{equation_posterior}
\end{align}
where $|\mathcal{C}|$ is the number of classes which is two here.
The $f_1(x)$ and $\pi_1$ are the \textit{likelihood (class conditional)} and \textit{prior} probabilities, respectively, and the denominator is the marginal probability. 

Therefore, Eq. (\ref{equation_posteriors_equal}) becomes:
\begin{align}
& \frac{f_1(x)\, \pi_1}{\sum_{i=1}^{|\mathcal{C}|} \mathbb{P}(X=x \,|\, x \in \mathcal{C}_i)\, \pi_i} \nonumber \\
& ~~~~~~~~~~~~ \overset{\text{set}}{=} \frac{f_2(x)\, \pi_2}{\sum_{i=1}^{|\mathcal{C}|} \mathbb{P}(X=x \,|\, x \in \mathcal{C}_i)\, \pi_i} \nonumber \\
& \implies f_1(x)\, \pi_1 = f_2(x)\, \pi_2. \label{equation_boundary_first_equation_2}
\end{align}

Now let us think of data as \textit{multivariate} data with dimensionality $d$.
The PDF for multivariate Gaussian distribution, $\b{x} \sim \mathcal{N}(\b{\mu}, \b{\Sigma})$ is:
\begin{align}\label{equation_multivariate_Gaussian}
f(\b{x}) = \frac{1}{\sqrt{(2\pi)^d |\b{\Sigma}|}} \exp\Big(\!\!- \frac{(\b{x} - \b{\mu})^\top \b{\Sigma}^{-1} (\b{x} - \b{\mu})}{2}\Big),
\end{align}
where $\b{x} \in \mathbb{R}^d$, $\b{\mu} \in \mathbb{R}^d$ is the mean, $\b{\Sigma} \in \mathbb{R}^{d \times d}$ is the covariance matrix, and $|.|$ is the determinant of matrix. The $\pi \approx 3.14$ in this equation should not be confused with the $\pi_k$ (prior) in Eq. (\ref{equation_boundary_first_equation_1}) or (\ref{equation_boundary_first_equation_2}).
Therefore, the Eq. (\ref{equation_boundary_first_equation_1}) or (\ref{equation_boundary_first_equation_2}) becomes:
\begin{equation}\label{equation_boundary_first_equation_Gaussian}
\begin{aligned}
&\frac{1}{\sqrt{(2\pi)^d |\b{\Sigma}_1|}} \exp\Big(\!\!- \frac{(\b{x} - \b{\mu}_1)^\top \b{\Sigma}_1^{-1} (\b{x} - \b{\mu}_1)}{2}\Big)\, \pi_1 \\
&= \frac{1}{\sqrt{(2\pi)^d |\b{\Sigma}_2|}} \exp\Big(\!\!- \frac{(\b{x} - \b{\mu}_2)^\top \b{\Sigma}_2^{-1} (\b{x} - \b{\mu}_2)}{2}\Big)\, \pi_2,
\end{aligned}
\end{equation}
where the distributions of the first and second class are $\mathcal{N}(\b{\mu}_1, \b{\Sigma}_1)$ and $\mathcal{N}(\b{\mu}_2, \b{\Sigma}_2)$, respectively.

\section{Linear Discriminant Analysis for Binary Classification}

In Linear Discriminant Analysis (LDA), we assume that the two classes have equal covariance matrices:
\begin{align}\label{equation_LDA_2classes_euqal_covariances}
\b{\Sigma}_1 = \b{\Sigma}_2 = \b{\Sigma}.
\end{align}
Therefore, the Eq. (\ref{equation_boundary_first_equation_Gaussian}) becomes:
\begin{align*}
&\frac{1}{\sqrt{(2\pi)^d |\b{\Sigma}|}} \exp\Big(\!\!- \frac{(\b{x} - \b{\mu}_1)^\top \b{\Sigma}^{-1} (\b{x} - \b{\mu}_1)}{2}\Big)\, \pi_1 \\
&= \frac{1}{\sqrt{(2\pi)^d |\b{\Sigma}|}} \exp\Big(\!\!- \frac{(\b{x} - \b{\mu}_2)^\top \b{\Sigma}^{-1} (\b{x} - \b{\mu}_2)}{2}\Big)\, \pi_2, \\
&\implies \exp\Big(\!\!- \frac{(\b{x} - \b{\mu}_1)^\top \b{\Sigma}^{-1} (\b{x} - \b{\mu}_1)}{2}\Big)\, \pi_1 \\
&~~~~~~~~~~~ = \exp\Big(\!\!- \frac{(\b{x} - \b{\mu}_2)^\top \b{\Sigma}^{-1} (\b{x} - \b{\mu}_2)}{2}\Big)\, \pi_2, \\
&\overset{(a)}{\implies} -\frac{1}{2} (\b{x} - \b{\mu}_1)^\top \b{\Sigma}^{-1} (\b{x} - \b{\mu}_1) + \ln(\pi_1) \\
&~~~~~~~~~~~ = -\frac{1}{2} (\b{x} - \b{\mu}_2)^\top \b{\Sigma}^{-1} (\b{x} - \b{\mu}_2) + \ln(\pi_2),
\end{align*}
where $(a)$ takes natural logarithm from the sides of equation.

We can simplify this term as:
\begin{align}
&(\b{x} - \b{\mu}_1)^\top \b{\Sigma}^{-1} (\b{x} - \b{\mu}_1) = (\b{x}^\top - \b{\mu}_1^\top) \b{\Sigma}^{-1} (\b{x} - \b{\mu}_1) \nonumber \\ 
&= \b{x}^\top \b{\Sigma}^{-1} \b{x} - \b{x}^\top \b{\Sigma}^{-1} \b{\mu}_1 - \b{\mu}_1^\top \b{\Sigma}^{-1} \b{x} + \b{\mu}_1^\top \b{\Sigma}^{-1} \b{\mu}_1 \nonumber \\
&\overset{(a)}{=} \b{x}^\top \b{\Sigma}^{-1} \b{x} + \b{\mu}_1^\top \b{\Sigma}^{-1} \b{\mu}_1 - 2\,\b{\mu}_1^\top \b{\Sigma}^{-1} \b{x}, \label{equation_simplified_term_1}
\end{align}
where $(a)$ is because $\b{x}^\top \b{\Sigma}^{-1} \b{\mu}_1 = \b{\mu}_1^\top \b{\Sigma}^{-1} \b{x}$ as it is a scalar and $\b{\Sigma}^{-1}$ is symmetric so $\b{\Sigma}^{-\top} = \b{\Sigma}^{-1}$.
Thus, we have:
\begin{align*}
& -\frac{1}{2} \b{x}^\top \b{\Sigma}^{-1} \b{x} -\frac{1}{2} \b{\mu}_1^\top \b{\Sigma}^{-1} \b{\mu}_1 + \b{\mu}_1^\top \b{\Sigma}^{-1} \b{x} + \ln(\pi_1) \\
& = -\frac{1}{2} \b{x}^\top \b{\Sigma}^{-1} \b{x} -\frac{1}{2} \b{\mu}_2^\top \b{\Sigma}^{-1} \b{\mu}_2 + \b{\mu}_2^\top \b{\Sigma}^{-1} \b{x} + \ln(\pi_2).
\end{align*}
Therefore, if we multiply the sides of equation by $2$, we have:
\begin{equation}\label{equation_LDA}
\begin{aligned}
&2\,\big(\b{\Sigma}^{-1} (\b{\mu}_2 - \b{\mu}_1)\big)^\top \b{x} \\
&~~~~~ + \b{\mu}_1^{\top} \b{\Sigma}^{-1} \b{\mu}_1 - \b{\mu}_2^{\top} \b{\Sigma}^{-1} \b{\mu}_2 + 2\,\ln(\frac{\pi_2}{\pi_1}) = 0,
\end{aligned}
\end{equation}
which is the equation of a line in the form of $\b{a}^\top \b{x} + b = 0$.
Therefore, if we consider Gaussian distributions for the two classes where the covariance matrices are assumed to be equal, the decision boundary of classification is a line. Because of linearity of the decision boundary which discriminates the two classes, this method is named \textit{linear} \textit{discriminant} analysis.

For obtaining Eq. (\ref{equation_LDA}), we brought the expressions to the right side which was corresponding to the second class; therefore, if we use $\delta(\b{x}): \mathbb{R}^d \rightarrow \mathbb{R}$ as the left-hand-side expression (function) in Eq. (\ref{equation_LDA}):
\begin{equation}\label{equation_LDA_2}
\begin{aligned}
&\delta(\b{x}) := 2\,\big(\b{\Sigma}^{-1} (\b{\mu}_2 - \b{\mu}_1)\big)^\top \b{x} \\
&~~~~~ + \b{\mu}_1^{\top} \b{\Sigma}^{-1} \b{\mu}_1 - \b{\mu}_2^{\top} \b{\Sigma}^{-1} \b{\mu}_2 + 2\,\ln(\frac{\pi_2}{\pi_1}),
\end{aligned}
\end{equation}
the class of an instance $\b{x}$ is estimated as:
\begin{align}\label{equation_classification_binary}
\widehat{\mathcal{C}}(x) = 
\left\{
\begin{array}{ll}
  1, & \text{if } \delta(\b{x}) < 0, \\
  2, & \text{if } \delta(\b{x}) > 0.
\end{array}
\right.
\end{align}

If the priors of two classes are equal, i.e., $\pi_1 = \pi_2$, the Eq. (\ref{equation_LDA}) becomes:
\begin{equation}\label{equation_LDA_equal_priors}
\begin{aligned}
&2\,\big(\b{\Sigma}^{-1} (\b{\mu}_2 - \b{\mu}_1)\big)^\top \b{x} \\
&~~~~~ + \b{\mu}_1^{\top} \b{\Sigma}^{-1} \b{\mu}_1 - \b{\mu}_2^{\top} \b{\Sigma}^{-1} \b{\mu}_2 = 0,
\end{aligned}
\end{equation}
whose left-hand-side expression can be considered as $\delta(\b{x})$ in Eq. (\ref{equation_classification_binary}).

\section{Quadratic Discriminant Analysis for Binary Classification}

In Quadratic Discriminant Analysis (QDA), we relax the assumption of equality of the covariance matrices:
\begin{align}
\b{\Sigma}_1 \neq \b{\Sigma}_2,
\end{align}
which means the covariances are not \textit{necessarily} equal (if they are actually equal, the decision boundary will be linear and QDA reduces to LDA).

Therefore, the Eq. (\ref{equation_boundary_first_equation_Gaussian}) becomes:
\begin{align*}
&\frac{1}{\sqrt{(2\pi)^d |\b{\Sigma}_1|}} \exp\Big(\!\!- \frac{(\b{x} - \b{\mu}_1)^\top \b{\Sigma}_1^{-1} (\b{x} - \b{\mu}_1)}{2}\Big)\, \pi_1 \\
&= \frac{1}{\sqrt{(2\pi)^d |\b{\Sigma}_2|}} \exp\Big(\!\!- \frac{(\b{x} - \b{\mu}_2)^\top \b{\Sigma}_2^{-1} (\b{x} - \b{\mu}_2)}{2}\Big)\, \pi_2, \\
&\overset{(a)}{\implies} -\frac{d}{2} \ln(2\pi) -\frac{1}{2} \ln(|\b{\Sigma}_1|) \\
&~~~~~~~~~~~ -\frac{1}{2} (\b{x} - \b{\mu}_1)^\top \b{\Sigma}_1^{-1} (\b{x} - \b{\mu}_1) + \ln(\pi_1) \\
&~~~~~~~~~~~ = -\frac{d}{2} \ln(2\pi) -\frac{1}{2} \ln(|\b{\Sigma}_2|) \\
&~~~~~~~~~~~ -\frac{1}{2} (\b{x} - \b{\mu}_2)^\top \b{\Sigma}_2^{-1} (\b{x} - \b{\mu}_2) + \ln(\pi_2),
\end{align*}
where $(a)$ takes natural logarithm from the sides of equation.
According to Eq. (\ref{equation_simplified_term_1}), we have:
\begin{align*}
& -\frac{1}{2} \ln(|\b{\Sigma}_1|) - \frac{1}{2} \b{x}^\top \b{\Sigma}_1^{-1} \b{x} -\frac{1}{2} \b{\mu}_1^\top \b{\Sigma}_1^{-1} \b{\mu}_1 \\
&+ \b{\mu}_1^\top \b{\Sigma}_1^{-1} \b{x} + \ln(\pi_1) \\
& = -\frac{1}{2} \ln(|\b{\Sigma}_2|) - \frac{1}{2} \b{x}^\top \b{\Sigma}_2^{-1} \b{x} -\frac{1}{2} \b{\mu}_2^\top \b{\Sigma}_2^{-1} \b{\mu}_2 \\
&+ \b{\mu}_2^\top \b{\Sigma}_2^{-1} \b{x} + \ln(\pi_2).
\end{align*}
Therefore, if we multiply the sides of equation by $2$, we have:
\begin{equation}\label{equation_QDA}
\begin{aligned}
&\b{x}^\top (\b{\Sigma}_1 - \b{\Sigma}_2)^{-1} \b{x} + 2\,(\b{\Sigma}_2^{-1} \b{\mu}_2 - \b{\Sigma}_1^{-1} \b{\mu}_1)^\top \b{x} \\
&+ (\b{\mu}_1^\top \b{\Sigma}_1^{-1} \b{\mu}_1 - \b{\mu}_2^\top \b{\Sigma}_2^{-1} \b{\mu}_2) + \ln\!\Big(\frac{|\b{\Sigma}_1|}{|\b{\Sigma}_2|}\Big) \\
&+ 2\,\ln(\frac{\pi_2}{\pi_1}) = 0,
\end{aligned}
\end{equation}
which is in the quadratic form $\b{x}^\top \b{A}\, \b{x} + \b{b}^\top \b{x} + c = 0$.
Therefore, if we consider Gaussian distributions for the two classes, the decision boundary of classification is quadratic. Because of quadratic decision boundary which discriminates the two classes, this method is named \textit{quadratic} \textit{discriminant} analysis.

For obtaining Eq. (\ref{equation_QDA}), we brought the expressions to the right side which was corresponding to the second class; therefore, if we use $\delta(\b{x}): \mathbb{R}^d \rightarrow \mathbb{R}$ as the left-hand-side expression (function) in Eq. (\ref{equation_QDA}):
\begin{equation}\label{equation_QDA_2}
\begin{aligned}
&\delta(\b{x}) := \b{x}^\top (\b{\Sigma}_1 - \b{\Sigma}_2)^{-1} \b{x} + 2\,(\b{\Sigma}_2^{-1} \b{\mu}_2 - \b{\Sigma}_1^{-1} \b{\mu}_1)^\top \b{x} \\
&+ (\b{\mu}_1^\top \b{\Sigma}_1^{-1} \b{\mu}_1 - \b{\mu}_2^\top \b{\Sigma}_2^{-1} \b{\mu}_2) + \ln\!\Big(\frac{|\b{\Sigma}_1|}{|\b{\Sigma}_2|}\Big) + 2\,\ln(\frac{\pi_2}{\pi_1}),
\end{aligned}
\end{equation}
the class of an instance $\b{x}$ is estimated as the Eq. (\ref{equation_classification_binary}).

If the priors of two classes are equal, i.e., $\pi_1 = \pi_2$, the Eq. (\ref{equation_LDA}) becomes:
\begin{equation}\label{equation_QDA_equal_priors}
\begin{aligned}
&\b{x}^\top (\b{\Sigma}_1 - \b{\Sigma}_2)^{-1} \b{x} + 2\,(\b{\Sigma}_2^{-1} \b{\mu}_2 - \b{\Sigma}_1^{-1} \b{\mu}_1)^\top \b{x} \\
&+ (\b{\mu}_1^\top \b{\Sigma}_1^{-1} \b{\mu}_1 - \b{\mu}_2^\top \b{\Sigma}_2^{-1} \b{\mu}_2) + \ln\!\Big(\frac{|\b{\Sigma}_1|}{|\b{\Sigma}_2|}\Big) = 0,
\end{aligned}
\end{equation}
whose left-hand-side expression can be considered as $\delta(\b{x})$ in Eq. (\ref{equation_classification_binary}).

\section{LDA and QDA for Multi-class Classification}

Now we consider multiple classes, which can be more than two, indexed by $k \in \{1,\dots,|\mathcal{C}|\}$.
Recall Eq. (\ref{equation_boundary_first_equation_1}) or (\ref{equation_boundary_first_equation_2}) where we are using the scaled posterior, i.e., $f_k(\b{x})\, \pi_k$. According to Eq. (\ref{equation_multivariate_Gaussian}), we have:
\begin{align*}
&f_k(\b{x})\, \pi_k \\
&= \frac{1}{\sqrt{(2\pi)^d |\b{\Sigma}_k|}} \exp\Big(\!\!- \frac{(\b{x} - \b{\mu}_k)^\top \b{\Sigma}_k^{-1} (\b{x} - \b{\mu}_k)}{2}\Big) \pi_k.
\end{align*}
Taking natural logarithm gives:
\begin{align*}
\ln(f_k(\b{x})\, \pi_k) = &-\frac{d}{2} \ln(2\pi) - \frac{1}{2} \ln(|\b{\Sigma}_k|) \\
&- \frac{1}{2} (\b{x} - \b{\mu}_k)^\top \b{\Sigma}_k^{-1} (\b{x} - \b{\mu}_k) + \ln(\pi_k).
\end{align*}
We drop the constant term $-(d/2) \ln(2\pi)$ which is the same for all classes (note that this term is multiplied before taking the logarithm). Thus, the scaled posterior of the $k$-th class becomes:
\begin{equation}\label{QDA_multiple_class_delta}
\begin{aligned}
\delta_k(\b{x}) := &- \frac{1}{2} \ln(|\b{\Sigma}_k|) \\
&- \frac{1}{2} (\b{x} - \b{\mu}_k)^\top \b{\Sigma}_k^{-1} (\b{x} - \b{\mu}_k) + \ln(\pi_k).
\end{aligned}
\end{equation}
In QDA, the class of the instance $\b{x}$ is estimated as:
\begin{align}\label{equation_max_delta}
\widehat{\mathcal{C}}(\b{x}) = \arg \max_k ~ \delta_k(\b{x}),
\end{align}
because it maximizes the posterior of that class. In this expression, $\delta_k(\b{x})$ is Eq. (\ref{QDA_multiple_class_delta}).

In LDA, we assume that the covariance matrices of the $k$ classes are equal:
\begin{align}
\b{\Sigma}_1 = \dots = \b{\Sigma}_{|\mathcal{C}|} = \b{\Sigma}.
\end{align}
Therefore, the Eq. (\ref{QDA_multiple_class_delta}) becomes:
\begin{align*}
&\delta_k(\b{x}) = - \frac{1}{2} \ln(|\b{\Sigma}|) \\
&- \frac{1}{2} (\b{x} - \b{\mu}_k)^\top \b{\Sigma}^{-1} (\b{x} - \b{\mu}_k) + \ln(\pi_k) = - \frac{1}{2} \ln(|\b{\Sigma}|) \\
&-\frac{1}{2} \b{x}^\top \b{\Sigma}^{-1} \b{x} - \frac{1}{2} \b{\mu}_k^\top \b{\Sigma}^{-1} \b{\mu}_k + \b{\mu}_k ^\top \b{\Sigma}^{-1} \b{x} + \ln(\pi_k).
\end{align*}
We drop the constant terms $- (1/2) \ln(|\b{\Sigma}|)$ and $- (1/2)\, \b{x}^\top \b{\Sigma}^{-1} \b{x}$ which are the same for all classes (note that before taking the logarithm, the term $- (1/2) \ln(|\b{\Sigma}|)$ is multiplied and the term $- (1/2)\, \b{x}^\top \b{\Sigma}^{-1} \b{x}$ is multiplied as an exponential term). Thus, the scaled posterior of the $k$-th class becomes: 
\begin{align}\label{LDA_multiple_class_delta}
\delta_k(\b{x}) := \b{\mu}_k ^\top \b{\Sigma}^{-1} \b{x} - \frac{1}{2} \b{\mu}_k^\top \b{\Sigma}^{-1} \b{\mu}_k + \ln(\pi_k).
\end{align}
In LDA, the class of the instance $\b{x}$ is determined by Eq. (\ref{equation_max_delta}), where $\delta(\b{x})$ is Eq. (\ref{LDA_multiple_class_delta}), because it maximizes the posterior of that class. 

In conclusion, QDA and LDA deal with maximizing the \textit{posterior} of classes but work with the \textit{likelihoods (class conditional)} and \textit{priors}.

\section{Estimation of Parameters in LDA and QDA}

In LDA and QDA, we have several parameters which are required in order to calculate the posteriors. These parameters are the means and the covariance matrices of classes and the priors of classes.

The priors of the classes are very tricky to calculate. It is somewhat a chicken and egg problem because we want to know the class probabilities (priors) to estimate the class of an instance but we do not have the priors and should estimate them.
Usually, the prior of the $k$-th class is estimated according to the sample size of the $k$-th class:
\begin{align}\label{equation_prior_Bernoulli}
\widehat{\pi}_k = \frac{n_k}{n},
\end{align}
where $n_k$ and $n$ are the number of training instances in the $k$-th class and in total, respectively.
This estimation considers Bernoulli distribution for choosing every instance out of the overall training set to be in the $k$-th class. 

The mean of the $k$-th class can be estimated using the Maximum Likelihood Estimation (MLE), or Method of Moments (MOM), for the mean of a Gaussian distribution:
\begin{align}\label{equation_LDA_mean_estimate}
\mathbb{R}^d \ni \widehat{\b{\mu}}_k = \frac{1}{n_k} \sum_{i=1}^{n} \b{x}_i\, \mathbb{I}\big(\mathcal{C}(\b{x}_i) = k\big),
\end{align}
where $\mathbb{I}(.)$ is the indicator function which is one and zero if its condition is satisfied and not satisfied, respectively. 

In QDA, the covariance matrix of the $k$-th class is estimated using MLE:
\begin{equation}
\begin{aligned}
&\mathbb{R}^{d \times d} \ni \widehat{\b{\Sigma}}_k = \\
&~~~~~~~~~~ \frac{1}{n_k} \sum_{i=1}^{n} (\b{x}_i - \widehat{\b{\mu}}_k) (\b{x}_i - \widehat{\b{\mu}}_k)^\top\, \mathbb{I}\big(\mathcal{C}(\b{x}_i) = k\big).
\end{aligned}
\end{equation}
Or we can use the \textit{unbiased} estimation of the covariance matrix:
\begin{equation}\label{equation_LDA_covariance_estimate}
\begin{aligned}
&\mathbb{R}^{d \times d} \ni \widehat{\b{\Sigma}}_k = \\
&~~~ \frac{1}{n_k - 1} \sum_{i=1}^{n} (\b{x}_i - \widehat{\b{\mu}}_k) (\b{x}_i - \widehat{\b{\mu}}_k)^\top\, \mathbb{I}\big(\mathcal{C}(\b{x}_i) = k\big).
\end{aligned}
\end{equation}

In LDA, we assume that the covariance matrices of the classes are equal; therefore, we use the weighted average of the estimated covariance matrices as the common covariance matrix in LDA:
\begin{align}\label{equation_LDA_covariance_estimate_unified}
\mathbb{R}^{d \times d} \ni \widehat{\b{\Sigma}} = \frac{\sum_{k=1}^{|\mathcal{C}|} n_k\, \widehat{\b{\Sigma}}_k}{\sum_{r=1}^{|\mathcal{C}|} n_r} = \frac{\sum_{k=1}^{|\mathcal{C}|} n_k\, \widehat{\b{\Sigma}}_k}{n},
\end{align}
where the weights are the cardinality of the classes.

\section{LDA and QDA are Metric Learning!}\label{section_metric_learning}

Recall Eq. (\ref{QDA_multiple_class_delta}) which is the scaled posterior for the QDA. 
First, assume that the covariance matrices are all equal (as we have in LDA) and they all are the identity matrix:
\begin{align}
\b{\Sigma}_1 = \dots = \b{\Sigma}_{|\mathcal{C}|} = \b{I},
\end{align}
which means that all the classes are assumed to be spherically distributed in the $d$ dimensional space.
After this assumption, the Eq. (\ref{QDA_multiple_class_delta}) becomes:
\begin{align}\label{eqaution_identityCovariances_differentPriors}
\delta_k(\b{x}) = &- \frac{1}{2} (\b{x} - \b{\mu}_k)^\top (\b{x} - \b{\mu}_k) + \ln(\pi_k),
\end{align}
because $|\b{I}| = 1$, $\ln(1)=0$, and $\b{I}^{-1} = \b{I}$.
If we assume that the priors are all equal, the term $\ln(\pi_k)$ is constant and can be dropped:
\begin{align}\label{equation_delta_and_Euclidean_distance}
\delta_k(\b{x}) = &- \frac{1}{2} (\b{x} - \b{\mu}_k)^\top (\b{x} - \b{\mu}_k) = -\frac{1}{2} d_k^2,
\end{align}
where $d_k$ is the Euclidean distance from the mean of the $k$-th class:
\begin{align}
d_k = ||\b{x} - \b{\mu}_k||_2 = \sqrt{(\b{x} - \b{\mu}_k)^\top (\b{x} - \b{\mu}_k)}.
\end{align}
Thus, the QDA or LDA reduce to simple Euclidean distance from the means of classes if the covariance matrices are all identity matrix and the priors are equal. Simple distance from the mean of classes is one of the simplest classification methods where the used metric is Euclidean distance.

\begin{figure}[!t]
\centering
\includegraphics[width=3in]{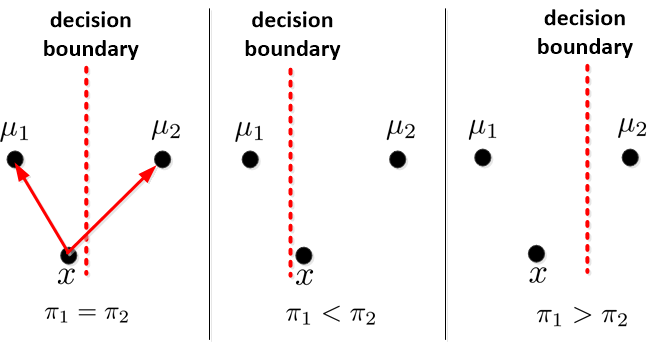}
\caption{The QDA and LDA where the covariance matrices are identity matrix. For equal priors, the QDA and LDA reduce to simple classification using Euclidean distance from means of classes. Changing the prior modifies the location of decision boundary where even one point can be classified differently for different priors.}
\label{figure_priors_metric}
\end{figure}

The Eq. (\ref{equation_delta_and_Euclidean_distance}) has a very interesting message. We know that in metric Multi-Dimensional Scaling (MDS) \cite{cox2000multidimensional} and kernel Principal Component Analysis (PCA), we have (see \cite{ham2004kernel} and Chapter 2 in \cite{strange2014open}):
\begin{align}\label{equation_kernel_and_distance}
\b{K} = - \frac{1}{2} \b{H} \b{D} \b{H},
\end{align}
where $\b{D} \in \mathbb{R}^{n \times n}$ is the distance matrix whose elements are the distances between the data instances, $\b{K} \in \mathbb{R}^{n \times n}$ is the kernel matrix over the data instances, $\mathbb{R}^{n \times n} \ni \b{H} := \b{I} - (1/n) \b{1}\b{1}^\top$ is the centering matrix, and $\mathbb{R}^n \ni \b{1} := [1, 1, \dots, 1]^\top$.
If the elements of the distance matrix $\b{D}$ are obtained using Euclidean distance, the MDS is equivalent to Principal Component Analysis (PCA) \cite{jolliffe2011principal}.

Comparing Eqs. (\ref{equation_delta_and_Euclidean_distance}) and (\ref{equation_kernel_and_distance}) shows an interesting connection between the posterior of a class in QDA and the kernel over the the data instances of the class. In this comparison, the Eq. (\ref{equation_kernel_and_distance}) should be considered for a class and not the entire data, so $\b{K} \in \mathbb{R}^{n_k \times n_k}$, $\b{D} \in \mathbb{R}^{n_k \times n_k}$, and $\b{H} \in \mathbb{R}^{n_k \times n_k}$.

Now, consider the case where still the covariance matrices are all identity matrix but the priors are not equal. In this case, we have Eq. (\ref{eqaution_identityCovariances_differentPriors}). If we take an exponential (inverse of logarithm) from this expression, the $\pi_k$ becomes a scale factor (weight). This means that we still are using distance metric to measure the distance of an instance from the means of classes but we are scaling the distances by the priors of classes. If a class happens more, i.e., its prior is larger, it must have a larger posterior so we reduce the distance from the mean of its class.
In other words, we move the decision boundary according to the prior of classes (see Fig. \ref{figure_priors_metric}).

As the next step, consider a more general case where the covariance matrices are not equal as we have in QDA. 
We apply Singular Value Decomposition (SVD) to the covariance matrix of the $k$-th class:
\begin{align*}
\b{\Sigma}_k = \b{U}_k\, \b{\Lambda}_k\, \b{U}_k^\top,
\end{align*}
where the left and right matrices of singular vectors are equal because the covariance matrix is symmetric. 
Therefore:
\begin{align*}
\b{\Sigma}_k^{-1} = \b{U}_k\, \b{\Lambda}_k^{-1}\, \b{U}_k^\top,
\end{align*}
where $\b{U}_k^{-1} = \b{U}_k^\top$ because it is an orthogonal matrix. 
Therefore, we can simplify the following term:
\begin{align*}
&(\b{x} - \b{\mu}_k)^\top \b{\Sigma}_k^{-1} (\b{x} - \b{\mu}_k) \\
&= (\b{x} - \b{\mu}_k)^\top \b{U}_k\, \b{\Lambda}_k^{-1}\, \b{U}_k^\top (\b{x} - \b{\mu}_k) \\
&= (\b{U}_k^\top \b{x} - \b{U}_k^\top \b{\mu}_k)^\top \b{\Lambda}_k^{-1}\, (\b{U}_k^\top \b{x} - \b{U}_k^\top \b{\mu}_k).
\end{align*}
As $\b{\Lambda}_k^{-1}$ is a diagonal matrix with non-negative elements (because it is covariance), we can decompose it as:
\begin{align*}
\b{\Lambda}_k^{-1} = \b{\Lambda}_k^{-1/2} \b{\Lambda}_k^{-1/2}.
\end{align*}
Therefore:
\begin{align*}
&(\b{U}_k^\top \b{x} - \b{U}_k^\top \b{\mu}_k)^\top \b{\Lambda}_k^{-1}\, (\b{U}_k^\top \b{x} - \b{U}_k^\top \b{\mu}_k) \\
&= (\b{U}_k^\top \b{x} - \b{U}_k^\top \b{\mu}_k)^\top \b{\Lambda}_k^{-1/2} \b{\Lambda}_k^{-1/2}\, (\b{U}_k^\top \b{x} - \b{U}_k^\top \b{\mu}_k) \\
&\overset{(a)}{=} (\b{\Lambda}_k^{-1/2} \b{U}_k^\top \b{x} - \b{\Lambda}_k^{-1/2} \b{U}_k^\top \b{\mu}_k)^\top  \\
&~~~~~ (\b{\Lambda}_k^{-1/2} \b{U}_k^\top \b{x} - \b{\Lambda}_k^{-1/2} \b{U}_k^\top \b{\mu}_k),
\end{align*}
where $(a)$ is because $\b{\Lambda}_k^{-\top/2} = \b{\Lambda}_k^{-1/2}$ because it is diagonal.
We define the following transformation:
\begin{align}
\phi_k: \b{x} \mapsto \b{\Lambda}_k^{-1/2} \b{U}_k^\top \b{x},
\end{align}
which also results in the transformation of the mean: $\phi_k: \b{\mu} \mapsto \b{\Lambda}_k^{-1/2} \b{U}_k^\top \b{\mu}$.
Therefore, the Eq. (\ref{QDA_multiple_class_delta}) can be restated as:
\begin{equation}
\begin{aligned}
&\delta_k(\b{x}) = - \frac{1}{2} \ln(|\b{\Sigma}_k|) \\
&- \frac{1}{2} \big(\phi_k(\b{x}) - \phi_k(\b{\mu}_k)\big)^\top \big(\phi_k(\b{x}) - \phi_k(\b{\mu}_k)\big) + \ln(\pi_k).
\end{aligned}
\end{equation}
Ignoring the terms $- (1/2) \ln(|\b{\Sigma}_k|)$ and $\ln(\pi_k)$, we can see that the transformation has changed the covariance matrix of the class to identity matrix. Therefore, the QDA (and also LDA) can be seen as simple comparison of distances from the means of classes after applying a transformation to the data of every class. In other words, we are learning the metric using the SVD of covariance matrix of every class. Thus, LDA and QDA can be seen as \textit{metric learning} \cite{yang2006distance,kulis2013metric} in a perspective.
Note that in metric learning, a valid distance metric is defined as \cite{yang2006distance}:
\begin{align}\label{equation_metric}
d_{\b{A}}^2(\b{x}, \b{\mu}_k) := ||\b{x} - \b{\mu}_k||_{\b{A}}^2 = (\b{x} - \b{\mu}_k)^\top \b{A}\, (\b{x} - \b{\mu}_k),
\end{align}
where $\b{A}$ is a positive semi-definite matrix, i.e., $\b{A} \succeq 0$.
In QDA, we are also using $(\b{x} - \b{\mu}_k)^\top \b{\Sigma}_k^{-1} (\b{x} - \b{\mu}_k)$. The covariance matrix is positive semi-definite according to the characteristics of covariance matrix. Moreover, according to characteristics of a positive semi-definite matrix, the inverse of a positive semi-definite matrix is positive semi-definite so $\b{\Sigma}_k^{-1} \succeq 0$. Therefore, QDA is using metric learning (and as will be discussed in next section, it can be seen as a \textit{manifold learning} method, too).

It is also noteworthy that the QDA and LDA can also be seen as \textit{Mahalanobis distance} \cite{mclachlan1999mahalanobis,de2000mahalanobis} which is also a metric:
\begin{align}\label{equation_Mahalanobis}
d_{M}^2(\b{x}, \b{\mu}) := ||\b{x} - \b{\mu}||_{M}^2 = (\b{x} - \b{\mu})^\top \b{\Sigma}^{-1} (\b{x} - \b{\mu}),
\end{align}
where $\b{\Sigma}$ is the covariance matrix of the cloud of data whose mean is $\b{\mu}$. The intuition of Mahalanobis distance is that if we have several data clouds (e.g., classes), the distance from the class with larger variance should be scaled down because that class is taking more of the space so it is more probable to happen. The scaling down shows in the inverse of covariance matrix. Comparing $(\b{x} - \b{\mu}_k)^\top \b{\Sigma}_k^{-1} (\b{x} - \b{\mu}_k)$ in QDA or LDA with Eq. (\ref{equation_Mahalanobis}) shows that QDA and LDA are sort of using Mahalanobis distance.

\section{LDA $\overset{?}{\equiv}$ FDA}\label{section_FDA}

In the previous section, we saw that LDA and QDA can be seen as metric learning. We know that metric learning can be seen as a family of manifold learning methods. We briefly explain the reason of this assertion: As $\b{A} \succeq 0$, we can say $\b{A} = \b{U}\b{U}^\top \succeq 0$. Therefore, Eq. (\ref{equation_metric}) becomes:
\begin{align*}
||\b{x} - \b{\mu}_k||_{\b{A}}^2 &= (\b{x} - \b{\mu}_k)^\top \b{U}\b{U}^\top\, (\b{x} - \b{\mu}_k) \\
&= (\b{U}^\top \b{x} - \b{U}^\top \b{\mu}_k)^\top (\b{U}^\top\b{x} - \b{U}^\top\b{\mu}_k),
\end{align*}
which means that metric learning can be seen as comparison of simple Euclidean distances after the transformation $\phi: \b{x} \mapsto \b{U}^\top \b{x}$ which is a projection into a subspace with projection matrix $\b{U}$. Thus, metric learning is a manifold learning approach. 
This gives a hint that the Fisher Discriminant Analysis (FDA) \cite{fisher1936use,welling2005fisher}, which is a manifold learning approach \cite{tharwat2017linear}, might have a connection to LDA; especially, because the names FDA and LDA are often used interchangeably in the literature. 
Actually, other names of FDA are Fisher LDA (FLDA) and even LDA.

We know that if we project (transform) the data of a class using a projection vector $\b{u} \in \mathbb{R}^p$ to a $p$ dimensional subspace ($p \leq d$), i.e.: 
\begin{align}\label{equation_FDA_projection}
\b{x} \mapsto \b{u}^\top \b{x},
\end{align}
for all data instances of the class, the mean and the covariance matrix of the class are transformed as:
\begin{align}
& \b{\mu} \mapsto \b{u}^\top \b{\mu}, \\
& \b{\Sigma} \mapsto \b{u}^\top \b{\Sigma}\, \b{u},
\end{align}
because of characteristics of mean and variance. 

The Fisher criterion \cite{xu2006analysis} is the ratio of the between-class variance, $\sigma^2_b$, and within-class variance, $\sigma^2_w$:
\begin{align}
f := \frac{\sigma^2_b}{\sigma^2_w} = \frac{(\b{u}^\top \b{\mu}_2 - \b{u}^\top \b{\mu}_1)^2}{\b{u}^\top \b{\Sigma}_2\, \b{u} + \b{u}^\top \b{\Sigma}_1\, \b{u}} = \frac{\big(\b{u}^\top (\b{\mu}_2 - \b{\mu}_1)\big)^2}{\b{u}^\top (\b{\Sigma}_2 + \b{\Sigma}_1)\, \b{u}}.
\end{align}
The FDA maximizes the Fisher criterion:
\begin{equation}\label{equation_Fisher_optimization}
\begin{aligned}
& \underset{\b{u}}{\text{maximize}}
& & \frac{\big(\b{u}^\top (\b{\mu}_2 - \b{\mu}_1)\big)^2}{\b{u}^\top (\b{\Sigma}_2 + \b{\Sigma}_1)\, \b{u}}, \\
\end{aligned}
\end{equation}
which can be restated as:
\begin{equation}\label{equation_Fisher_optimization_2}
\begin{aligned}
& \underset{\b{u}}{\text{maximize}}
& & \big(\b{u}^\top (\b{\mu}_2 - \b{\mu}_1)\big)^2, \\
& \text{subject to}
& & \b{u}^\top (\b{\Sigma}_2 + \b{\Sigma}_1)\, \b{u} = 1,
\end{aligned}
\end{equation}
according to Rayleigh-Ritz quotient method \cite{croot2005rayleigh}.
The Lagrangian \cite{boyd2004convex} is: 
\begin{align*}
\mathcal{L} = \big(\b{u}^\top (\b{\mu}_2 - \b{\mu}_1)\big)^2 - \lambda \big(\b{u}^\top (\b{\Sigma}_2 + \b{\Sigma}_1)\, \b{u} - 1\big),
\end{align*}
where $\lambda$ is the Lagrange multiplier. Equating the derivative of $\mathcal{L}$ to zero gives: 
\begin{align*}
&\frac{\partial \mathcal{L}}{\partial \b{u}} = 2\,(\b{\mu}_2 - \b{\mu}_1) (\b{\mu}_2 - \b{\mu}_1)^\top \b{u} - 2\,\lambda\, (\b{\Sigma}_2 + \b{\Sigma}_1)\, \b{u} \overset{\text{set}}{=} \b{0} \\ 
& \implies (\b{\mu}_2 - \b{\mu}_1) (\b{\mu}_2 - \b{\mu}_1)^\top \b{u} = \lambda\, (\b{\Sigma}_2 + \b{\Sigma}_1)\, \b{u},
\end{align*}
which is a generalized eigenvalue problem $\big((\b{\mu}_2 - \b{\mu}_1) (\b{\mu}_2 - \b{\mu}_1)^\top, (\b{\Sigma}_2 + \b{\Sigma}_1)\big)$ according to \cite{ghojogh2019eigenvalue}. The projection vector is the eigenvector of $(\b{\Sigma}_2 + \b{\Sigma}_1)^{-1} (\b{\mu}_2 - \b{\mu}_1) (\b{\mu}_2 - \b{\mu}_1)^\top$; therefore, we can say:
\begin{align}\label{equation_u_LDA_and_FDA}
\b{u} \propto (\b{\Sigma}_2 + \b{\Sigma}_1)^{-1} (\b{\mu}_2 - \b{\mu}_1) (\b{\mu}_2 - \b{\mu}_1)^\top.
\end{align}
In LDA, the equality of covariance matrices is assumed. Thus, according to Eq. (\ref{equation_LDA_2classes_euqal_covariances}), we can say:
\begin{align}
\b{u} &\propto (2\,\b{\Sigma})^{-1} (\b{\mu}_2 - \b{\mu}_1) (\b{\mu}_2 - \b{\mu}_1)^\top \nonumber\\
&\propto \b{\Sigma}^{-1} (\b{\mu}_2 - \b{\mu}_1) (\b{\mu}_2 - \b{\mu}_1)^\top.
\end{align}
According to Eq. (\ref{equation_FDA_projection}), we have:
\begin{align}\label{equation_FDA_projection_2}
\b{u}^\top \b{x} \propto \big(\b{\Sigma}^{-1} (\b{\mu}_2 - \b{\mu}_1) (\b{\mu}_2 - \b{\mu}_1)^\top\big)^\top \b{x}.
\end{align}
Comparing Eq. (\ref{equation_FDA_projection_2}) with Eq. (\ref{equation_LDA_equal_priors}) shows that LDA and FDA are equivalent up to a scaling factor $\big(\b{\mu}_1 - \b{\mu}_2)^{\top} \b{\Sigma}^{-1} (\b{\mu}_1 - \b{\mu}_2)$ (note that this term is multiplied as an exponential factor before taking logarithm to obtain Eq. (\ref{equation_LDA_equal_priors}), so this term a scaling factor).
Hence, we can say:
\begin{align}
\text{LDA} \equiv \text{FDA}.
\end{align}
In other words, FDA projects into a subspace. On the other hand, according to Section \ref{section_metric_learning}, LDA can be seen as a metric learning with a subspace where the Euclidean distance is used after projecting onto that subspace. \textit{The two subspaces of FDA and LDA are the same subspace}\footnote{It should be noted that in manifold (subspace) learning, the scale does not matter because all the distances scale similarly.}. One might wonder how a linear classifier (LDA) can be equivalent to a linear dimensionality reduction method (FDA). The answer is that \textit{projection of data onto the FDA subspace and then classifying them by Euclidean distance in the subspace is equivalent to classification by the LDA classifier in the input space.}

Note that LDA assumes \textit{one} (and not several) Gaussian for every class and so does the FDA. That is why FDA faces problem for multi-modal data \cite{sugiyama2007dimensionality}.

\section{Relation to Logistic Regression}

According to Eqs. (\ref{equation_multivariate_Gaussian}) and (\ref{equation_prior_Bernoulli}), Gaussian and Bernoulli distributions are used for likelihood (class conditional) and prior, respectively, in LDA and QDA. 
Thus, we are making assumptions for the likelihood and prior, although we finally work with posterior in LDA and QDA according to Eq. (\ref{equation_boundary_first_equation_2}).
\textit{Logistic regression} \cite{kleinbaum2002logistic} says why do we make assumptions on the likelihood and prior when we want to work on posterior finally. Let us make assumption directly for the posterior. 

In logistic regression, first a linear function is applied to the data to have $\b{\beta}^\top \b{x}'$ where $\mathbb{R}^{d+1} \ni \b{x}' = [\b{x}^\top, 1]^\top$ and $\b{\beta} \in \mathbb{R}^{d+1}$ include the intercept. Then, logistic function is used in order to have a value in range $(0,1)$ to simulate probability. Therefore, in logistic regression, the posterior is assumed to be:
\begin{equation}\label{equation_logistic_regression_posterior}
\begin{aligned}
&\mathbb{P}(\mathcal{C}(\b{x}) \,|\, X = \b{x}) \\
&= \Big(\frac{\exp(\b{\beta}^\top \b{x}')}{1+\exp(\b{\beta}^\top \b{x}')}\Big)^{\mathcal{C}(\b{x})} \Big(\frac{1}{1+\exp(\b{\beta}^\top \b{x}')}\Big)^{1-\mathcal{C}(\b{x})},
\end{aligned}
\end{equation}
where $\mathcal{C}(\b{x}) \in \{-1, +1\}$ for the two classes.
Logistic regression considers the coefficient $\b{\beta}$ as the parameter to be optimized and uses Newton's method \cite{boyd2004convex} for the optimization. 
Therefore, in summary, logistic regression makes assumption on the posterior while LDA and QDA make assumption on likelihood and prior.

\section{Relation to Bayes Optimal Classifier and Gaussian Naive Bayes}

The Bayes classifier maximizes the posteriors of the classes \cite{murphy2012machine}:
\begin{align}\label{equation_bayes_classification_1}
\widehat{\mathcal{C}}(\b{x}) = \arg \max_k ~ \mathbb{P}(\b{x} \in \mathcal{C}_k \,|\, X=\b{x}).
\end{align}
According to Eq. (\ref{equation_posterior}) and Bayes rule, we have:
\begin{align}\label{equation_posterior_as_likelihood_and_prior}
\mathbb{P}(\b{x} \in \mathcal{C}_k \,|\, X=\b{x}) \propto \mathbb{P}(X=\b{x} \,|\, \b{x} \in \mathcal{C}_k)\, \underbrace{\mathbb{P}(\b{x} \in \mathcal{C}_k)}_{\pi_k},
\end{align}
where the denominator of posterior (the marginal) which is:
\begin{align*}
\mathbb{P}(X=\b{x}) = \sum_{r=1}^{|\mathcal{C}|} \mathbb{P}(X=\b{x} \,|\, \b{x} \in \mathcal{C}_r)\, \pi_r,
\end{align*}
is ignored because it is not dependent on the classes $\mathcal{C}_1$ to $\mathcal{C}_{|\mathcal{C}|}$.

According to Eq. (\ref{equation_posterior_as_likelihood_and_prior}), the posterior can be written in terms of likelihood and prior; therefore, Eq. (\ref{equation_bayes_classification_1}) can be restated as:
\begin{align}\label{equation_bayes_classification_2}
\widehat{\mathcal{C}}(\b{x}) = \arg \max_k ~ \pi_k\, \mathbb{P}(X=\b{x} \,|\, \b{x} \in \mathcal{C}_k).
\end{align}

Note that the Bayes classifier does not make any assumption on the posterior, prior, and likelihood, unlike LDA and QDA which assume the uni-modal Gaussian distribution for the likelihood (and we \textit{may} assume Bernoulli distribution for the prior in LDA and QDA according to Eq. (\ref{equation_prior_Bernoulli})).
Therefore, we can say the difference of Bayes and QDA is in assumption of \textit{uni-modal} Gaussian distribution for the likelihood (class conditional); hence, if the likelihoods are already uni-modal Gaussian, the Bayes classifier reduces to QDA.
Likewise, the difference of Bayes and LDA is in assumption of Gaussian distribution for the likelihood (class conditional) and equality of covariance matrices of classes; thus, if the likelihoods are already Gaussian and the covariance matrices are already equal, the Bayes classifier reduces to LDA.

It is noteworthy that the Bayes classifier is an optimal classifier because it can be seen as an ensemble of hypotheses (models) in the hypothesis (model) space and no other ensemble of hypotheses can outperform it (see Chapter 6, Page 175 in \cite{mitchell1997machine}). In the literature, it is referred to as \textit{Bayes optimal classifier}. 
To better formulate the explained statements, the Bayes optimal classifier estimates the class as:
\begin{align}\label{equation_bayes_optimal_classification_1}
\widehat{\mathcal{C}}(\b{x}) = \arg \max_{\mathcal{C}_k \in \mathcal{C}} ~ \sum_{h_j \in \mathcal{H}} \mathbb{P}(\mathcal{C}_k \,|\, h_j)\, \mathbb{P}(\mathcal{D} \,|\, h_j)\, \mathbb{P}(h_j),
\end{align}
where $\mathcal{C} := \{\mathcal{C}_1, \dots, \mathcal{C}_{|\mathcal{C}|}\}$, $\mathcal{D} := \{\b{x}_i\}_{i=1}^n$ is the training set, $h_j$ is a hypothesis for estimating the class of instances, and $\mathcal{H}$ is the hypothesis space including all possible hypotheses. 

According to Bayes rule, similar to what we had for Eq. (\ref{equation_posterior_as_likelihood_and_prior}), we have:
\begin{align*}
\mathbb{P}(h_j \,|\, \mathcal{D}) \propto \mathbb{P}(\mathcal{D} \,|\, h_j)\, \mathbb{P}(h_j).
\end{align*}
Therefore, Eq. (\ref{equation_bayes_optimal_classification_1}) becomes \cite{mitchell1997machine}:
\begin{align}\label{equation_bayes_optimal_classification_2}
\widehat{\mathcal{C}}(\b{x}) = \arg \max_{\mathcal{C}_k \in \mathcal{C}} ~ \sum_{h_j \in \mathcal{H}} \mathbb{P}(\mathcal{C}_k \,|\, h_j)\, \mathbb{P}(h_j \,|\, \mathcal{D}),
\end{align}

In conclusion, the Bayes classifier is optimal. Therefore, if the likelihoods of classes are Gaussian, QDA is an optimal classifier and if the likelihoods are Gaussian and the covariance matrices are equal, the LDA is an optimal classifier. 
Often, the distributions in the natural life are Gaussian; especially, because of central limit theorem \cite{hazewinkel2001central}, the summation of independent and identically distributed (iid) variables is Gaussian and the signals usually add in the real world.
This explains why LDA and QDA are very effective classifiers in machine learning. We also saw that FDA is equivalent to LDA. Thus, the reason of effectiveness of the powerful FDA classifier becomes clear. We have seen the very successful performance of FDA and LDA in different applications, such as face recognition \cite{belhumeur1997eigenfaces,etemad1997discriminant,zhao1999subspace}, action recognition \cite{ghojogh2017fisherposes,mokari2018recognizing}, and EEG classification \cite{malekmohammadi2019efficient}.

Implementing Bayes classifier is difficult in practice so we approximate it by \textit{naive Bayes} \cite{zhang2004optimality}.
If $x_j$ denotes the $j$-th dimension (feature) of $\b{x} = [x_1, \dots, x_d]^\top$, Eq. (\ref{equation_bayes_classification_2}) is restated as:
\begin{align}\label{equation_bayes_classification_3}
\widehat{\mathcal{C}}(\b{x}) = \arg \max_k ~ \pi_k\, \mathbb{P}(x_1, x_2, \dots, x_d \,|\, \b{x} \in \mathcal{C}_k).
\end{align}
The term $\mathbb{P}(x_1, x_2, \dots, x_d \,|\, \b{x} \in \mathcal{C}_k)$ is very difficult to compute as the features are possibly correlated. Naive Bayes relaxes this possibility and naively assumes that the features are conditionally independent ($\indep$) when they are conditioned on the class:
\begin{equation*}
\begin{aligned}
&\mathbb{P}(x_1, x_2, \dots, x_d \,|\, \b{x} \in \mathcal{C}_k) \\
&\overset{\indep}{\approx} \mathbb{P}(x_1 \,|\, \mathcal{C}_k)\, \mathbb{P}(x_2 \,|\, \mathcal{C}_k)\, \cdots\, \mathbb{P}(x_d \,|\, \mathcal{C}_k) = \prod_{j=1}^d \mathbb{P}(x_j \,|\, \mathcal{C}_k).
\end{aligned}
\end{equation*}
Therefore, Eq. (\ref{equation_bayes_classification_3}) becomes:
\begin{align}\label{equation_naive_bayes_classification}
\widehat{\mathcal{C}}(\b{x}) = \arg \max_k ~ \pi_k\, \prod_{j=1}^d \mathbb{P}(x_j \,|\, \mathcal{C}_k).
\end{align}
In \textit{Gaussian naive Bayes}, univariate Gaussian distribution is assumed for the likelihood (class conditional) of every feature:
\begin{align}\label{equation_Gaussian_naive_bayes_prob}
\mathbb{P}(x_j \,|\, \mathcal{C}_k) = \frac{1}{\sqrt{2 \pi \sigma_k^2}} \exp\Big(\!\!-\frac{(x_j - \mu_k)^2}{2 \sigma_k^2}\Big),
\end{align}
where the mean and unbiased variance are estimated as:
\begin{align}
&\mathbb{R} \ni \widehat{\mu}_k = \frac{1}{n_k} \sum_{i=1}^{n} x_{i,j}\, \mathbb{I}\big(\mathcal{C}(\b{x}_i) = k\big), \label{equation_naiveBayes_mean_estimate} \\
&\mathbb{R} \ni \widehat{\sigma}_k^2 = \frac{1}{n_k - 1} \sum_{i=1}^n (x_{i,j} - \widehat{\mu}_k)^2\, \mathbb{I}\big(\mathcal{C}(\b{x}_i) = k\big), \label{equation_naiveBayes_covariance_estimate}
\end{align}
where $x_{i,j}$ denotes the $j$-th feature of the $i$-th training instance.
The prior can again be estimated using Eq. (\ref{equation_prior_Bernoulli}).

According to Eqs. (\ref{equation_naive_bayes_classification}) and (\ref{equation_Gaussian_naive_bayes_prob}), Gaussian naive Bayes is equivalent to QDA where the covariance matrices are \textit{diagonal}, i.e., the off-diagonal of the covariance matrices are ignored. Therefore, we can say that QDA is more powerful than Gaussian naive Bayes because Gaussian naive Bayes is a simplified version of QDA.
Moreover, it is obvious that Gaussian naive Bayes and QDA are equivalent for \textit{one} dimensional data.
Comparing to LDA, the Gaussian naive Bayes is equivalent to LDA if the covariance matrices are diagonal and they are all equal, i.e., $\sigma_1^2 = \dots = \sigma_{|\mathcal{C}|}^2$; therefore, LDA and Gaussian naive Bayes have their own assumptions, one on the off-diagonal of covariance matrices and the other one on equality of the covariance matrices.
As Gaussian naive Bayes has some level of optimality \cite{zhang2004optimality}, it becomes clear why LDA and QDA are such effective classifiers.

\section{Relation to Likelihood Ratio Test}

Consider two hypotheses for estimating some parameter, a null hypothesis $H_0$ and an alternative hypothesis $H_A$.
The probability $\mathbb{P}(\text{reject } H_0 \,|\, H_0)$ is called type 1 error, false positive error, or false alarm error. The probability $\mathbb{P}(\text{accept } H_0 \,|\, H_A)$ is called type 2 error or false negative error. The $\mathbb{P}(\text{reject } H_0 \,|\, H_0)$ is also called \textit{significance level}, while $1 - \mathbb{P}(\text{accept } H_0 \,|\, H_A) = \mathbb{P}(\text{reject } H_0 \,|\, H_A)$ is called \textit{power}.

If $L(\theta_A)$ and $L(\theta_0)$ are the likelihoods (probabilities) for the alternative and null hypotheses, the likelihood ratio is:
\begin{align}
\Lambda = \frac{L(\theta_A)}{L(\theta_0)} = \frac{f(\b{x}; \theta_A)}{f(\b{x}; \theta_0)}.    
\end{align}
The Likelihood Ratio Test (LRT) \cite{casella2002statistical} rejects the $H_0$ in favor of $H_A$ if the likelihood ratio is greater than a threshold, i.e., $\Lambda \geq t$.
The LRT is a very effective statistical test because according to the Neyman-Pearson lemma \cite{neyman1933ix}, it has the largest power among all statistical tests with the same significance level. 

If the sample size is large, $n \rightarrow \infty$, and the $\theta_A$ is estimated using MLE, the logarithm of the likelihood ratio asymptotically has the distribution of $\chi^2$ under the null hypothesis \cite{white1984asymptotic,casella2002statistical}:
\begin{align}
2 \ln(\Lambda) \overset{H_0}{\sim} \chi_{(df)}^2,
\end{align}
where the degree of freedom of $\chi^2$ distribution is $df := \text{dim}(H_A) - \text{dim}(H_0)$ and $\text{dim}(.)$ is the number of unspecified parameters in the hypothesis.

There is a connection between LDA or QDA and the LRT \cite{lachenbruch1979discriminant}.
Recall Eq. (\ref{equation_boundary_first_equation_1}) or (\ref{equation_boundary_first_equation_2}) which can be restated as:
\begin{align}\label{equation_likelihood_ratio_QDA_1}
\frac{f_2(\b{x})\, \pi_2}{f_1(\b{x})\, \pi_1} = 1, 
\end{align}
which is for the decision boundary. 
The Eq. (\ref{equation_classification_binary}) dealt with the difference of $f_2(x)\, \pi_2$ and $f_1(x)\, \pi_1$; however, here we are dealing with their ratio. Recall Fig. \ref{figure_two_Gaussians} where if we move $x^*$ to the right and left, the ratio $f_2(x^*)\, \pi_2 / f_1(x^*)\, \pi_1$ decreases and increases, respectively, because the probabilities of the first and second class happening change. In other words, moving the $x^*$ changes the significance level and power.
Therefore, Eq. (\ref{equation_likelihood_ratio_QDA_1}) can be used to have a statistical test where the posteriors are used in the ratio, as we also used posteriors in LDA and QDA. 
The null/alternative hypothesis an be considered to be the mean and covariance of the first/second class. In other words, the two hypotheses say that the point belongs to a specific class.
Hence, if the ratio is larger than a value $t$, the instance $\b{x}$ is estimated to belong to the second class; otherwise, the first class is chosen.
According to Eq. (\ref{equation_multivariate_Gaussian}), the Eq. (\ref{equation_likelihood_ratio_QDA_1}) becomes:
\begin{align}\label{equation_likelihood_ratio_QDA_2}
\frac{(|\b{\Sigma}_2|)^{-1/2} \exp\big(\!-\frac{1}{2} (\b{x} - \b{\mu}_2)^\top \b{\Sigma}_2^{-1} (\b{x} - \b{\mu}_2)\big)\, \pi_2}{(|\b{\Sigma}_1|)^{-1/2} \exp\big(\!-\frac{1}{2} (\b{x} - \b{\mu}_1)^\top \b{\Sigma}_1^{-1} (\b{x} - \b{\mu}_1)\big)\, \pi_1} \geq t,
\end{align}
for QDA.
In LDA, the covariance matrices are equal, so:
\begin{align}\label{equation_likelihood_ratio_LDA}
\frac{\exp\big(\!-\frac{1}{2} (\b{x} - \b{\mu}_2)^\top \b{\Sigma}^{-1} (\b{x} - \b{\mu}_2)\big)\, \pi_2}{\exp\big(\!-\frac{1}{2} (\b{x} - \b{\mu}_1)^\top \b{\Sigma}^{-1} (\b{x} - \b{\mu}_1)\big)\, \pi_1} \geq t.
\end{align}
As can be seen, changing the priors change impacts the ratio as expected.
Moreover, the value of $t$ can be chosen according to the desired significance level in the $\chi^2$ distribution using the $\chi^2$ table.
The Eqs. (\ref{equation_likelihood_ratio_QDA_2}) and (\ref{equation_likelihood_ratio_LDA}) show the relation of LDA and QDA with LRT.
As the LRT has the largest power \cite{neyman1933ix}, the effectiveness of LDA and QDA in classification is explained from a hypothesis testing point of view.

\begin{figure*}[!t]
% https://stackoverflow.com/questions/41041534/side-by-side-subfigure-in-sharelatex
\centering
\subfigure[][]{%
\includegraphics[height=2.1in]{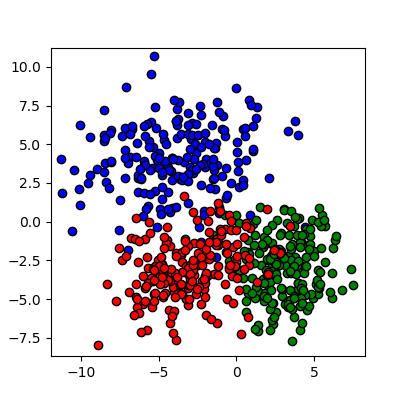}} 
\subfigure[][]{%
\includegraphics[height=2.1in]{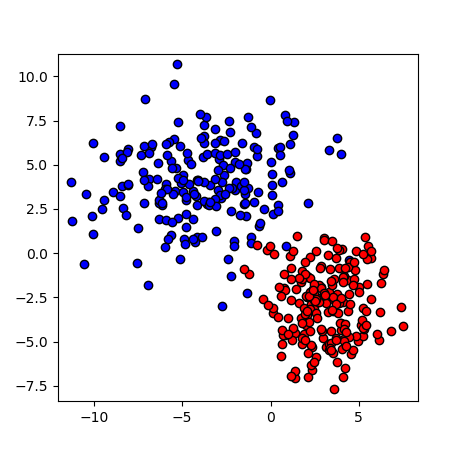}} 
\subfigure[][]{%
\includegraphics[height=2.1in]{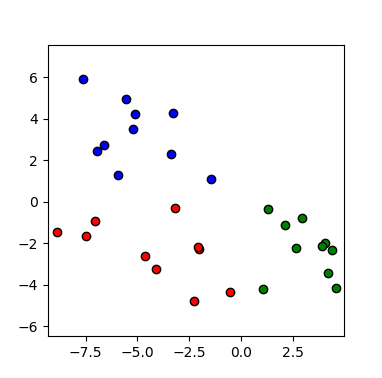}} 
\subfigure[][]{%
\includegraphics[height=2in]{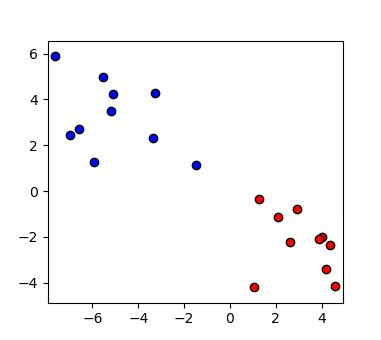}} 
\subfigure[][]{%
\includegraphics[height=2in]{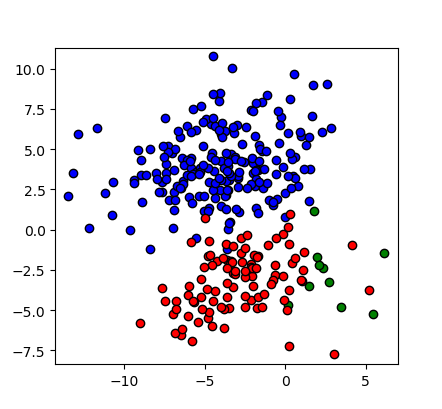}} 
\subfigure[][]{%
\includegraphics[height=2in]{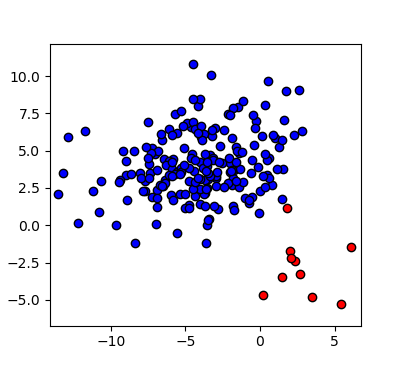}} 
\subfigure[][]{%
\includegraphics[height=2.1in]{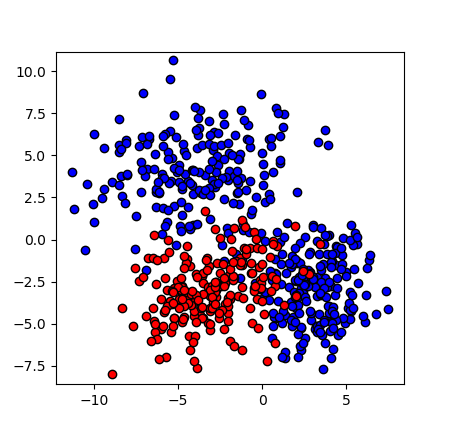}} 
\caption{The synthetic dataset: (a) three classes each with size $200$, (b) two classes each with size $200$, (c) three classes each with size $10$, (d) two classes each with size $10$, (e) three classes with sizes $200$, $100$, and $10$, (f) two classes with sizes $200$ and $10$, and (g) two classes with sizes $400$ and $200$ where the larger class has two modes.}
\label{figure_datasets}%
\end{figure*}

\begin{figure*}[!t]
% https://stackoverflow.com/questions/41041534/side-by-side-subfigure-in-sharelatex
\centering
\subfigure[][]{%
\includegraphics[height=2.1in]{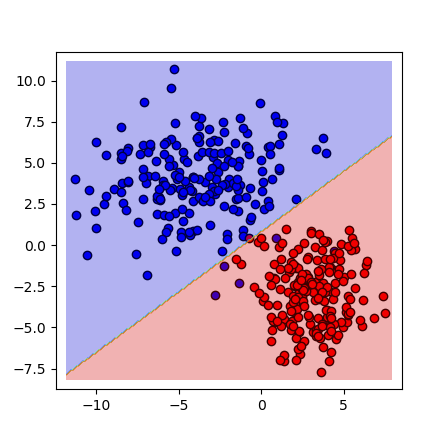}} 
\subfigure[][]{%
\includegraphics[height=2.1in]{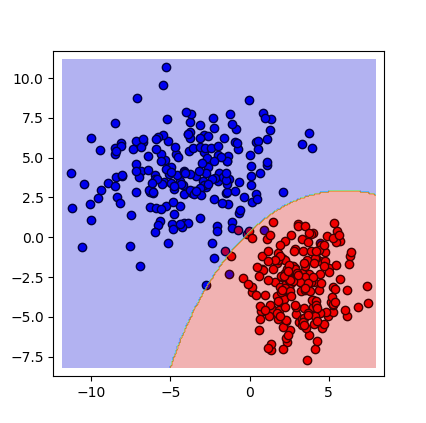}} 
\hspace{0pt}%
\subfigure[][]{%
\includegraphics[height=2.1in]{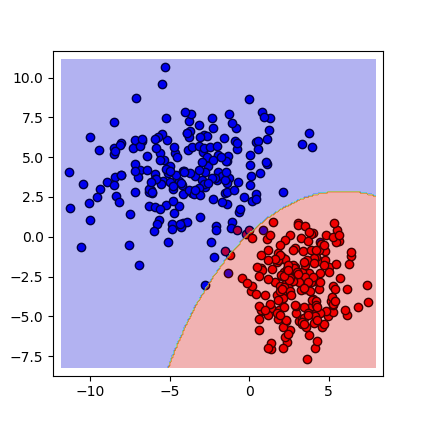}} \\
\subfigure[][]{%
\includegraphics[height=2.1in]{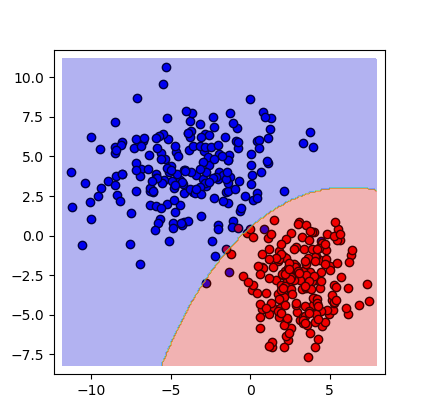}}%
\subfigure[][]{%
\includegraphics[height=2.1in]{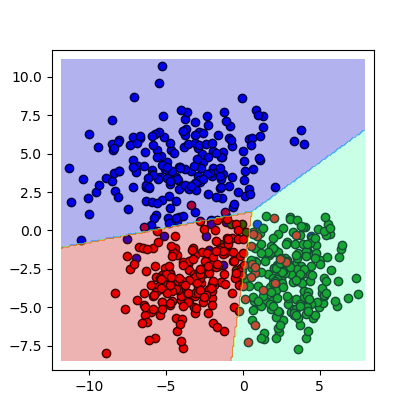}}%
\subfigure[][]{%
\includegraphics[height=2.1in]{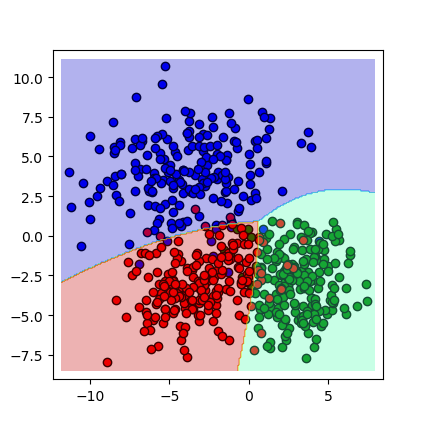}} \\
\subfigure[][]{%
\includegraphics[height=2.1in]{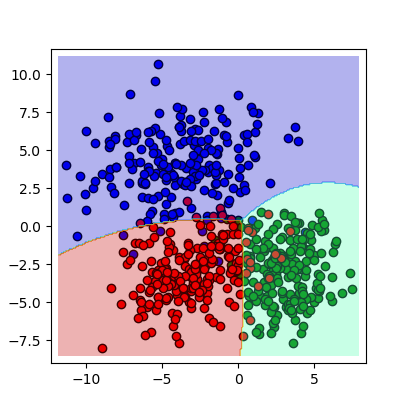}}
\subfigure[][]{%
\includegraphics[height=2.1in]{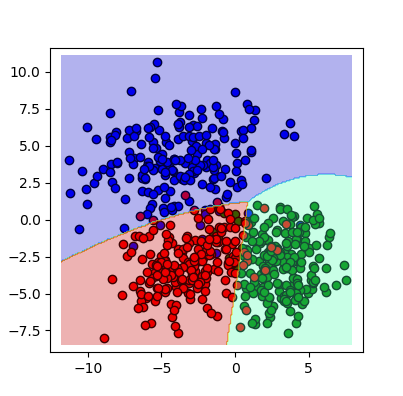}}%
\caption{Experiments with equal class sample sizes: (a) LDA for two classes, (b) QDA for two classes, (c) Gaussian naive Bayes for two classes, (d) Bayes for two classes, (e) LDA for three classes, (f) QDA for three classes, (g) Gaussian naive Bayes for three classes, and (h) Bayes for three classes.}%
\label{figure_equal_sizes}%
\end{figure*}

\begin{figure*}[!t]
% https://stackoverflow.com/questions/41041534/side-by-side-subfigure-in-sharelatex
\centering
\subfigure[][]{%
\includegraphics[height=2in]{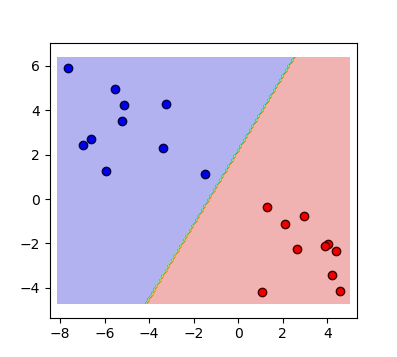}} 
\subfigure[][]{%
\includegraphics[height=1.95in]{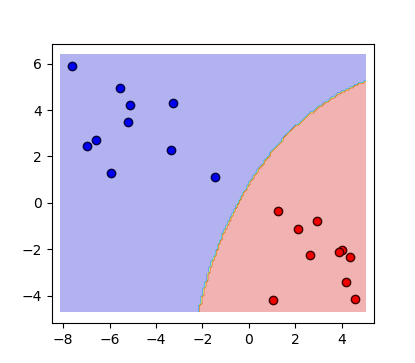}} 
\hspace{0pt}%
\subfigure[][]{%
\includegraphics[height=1.95in]{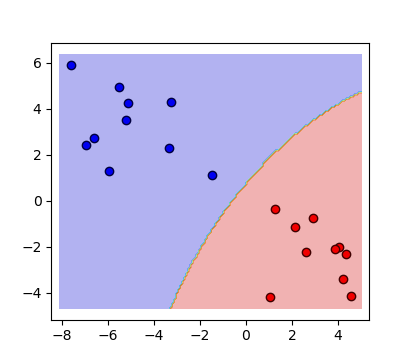}} \\
\subfigure[][]{%
\includegraphics[height=2in]{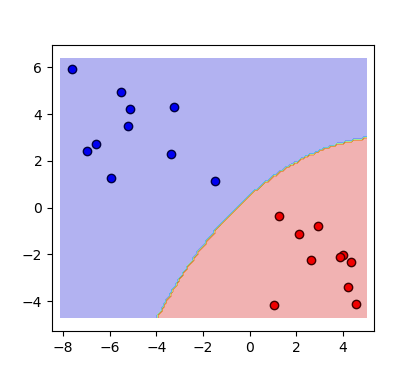}}%
\subfigure[][]{%
\includegraphics[height=2in]{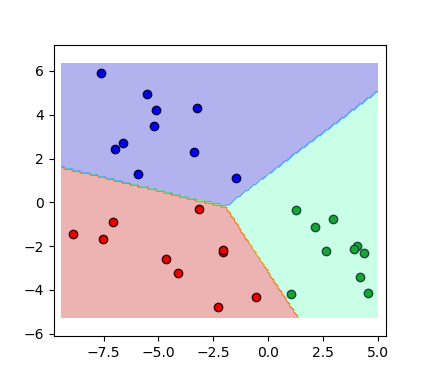}}%
\subfigure[][]{%
\includegraphics[height=2in]{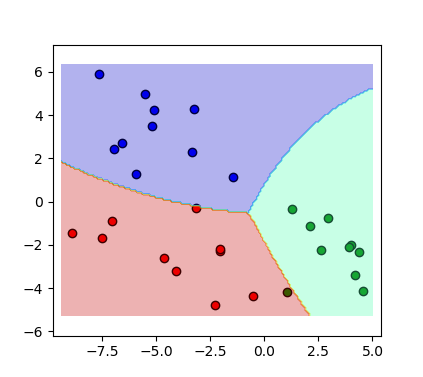}} \\
\subfigure[][]{%
\includegraphics[height=2in]{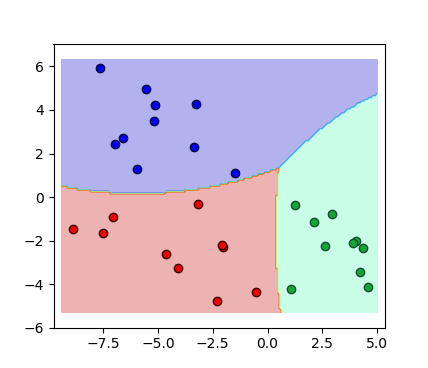}}
\subfigure[][]{%
\includegraphics[height=2in]{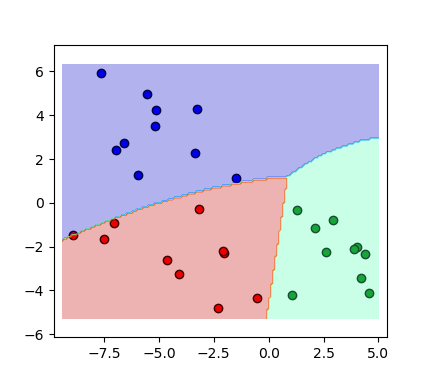}}%
\caption{Experiments with small class sample sizes: (a) LDA for two classes, (b) QDA for two classes, (c) Gaussian naive Bayes for two classes, (d) Bayes for two classes, (e) LDA for three classes, (f) QDA for three classes, (g) Gaussian naive Bayes for three classes, and (h) Bayes for three classes.}%
\label{figure_small_sizes}%
\end{figure*}

\begin{figure*}[!t]
% https://stackoverflow.com/questions/41041534/side-by-side-subfigure-in-sharelatex
\centering
\subfigure[][]{%
\includegraphics[height=1.9in]{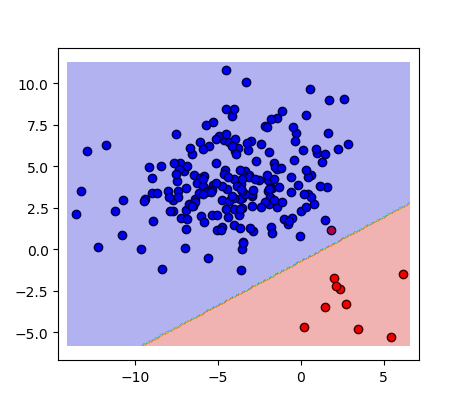}} 
\subfigure[][]{%
\includegraphics[height=1.9in]{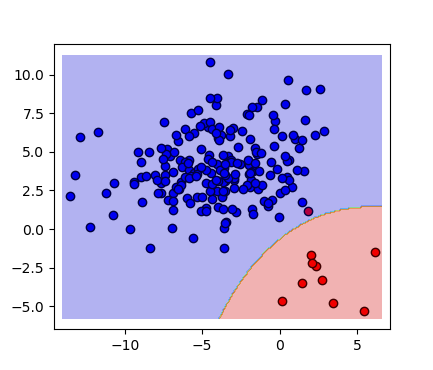}} 
\hspace{0pt}%
\subfigure[][]{%
\includegraphics[height=1.9in]{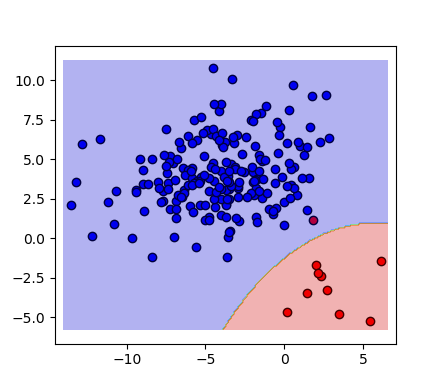}} \\
\subfigure[][]{%
\includegraphics[height=2in]{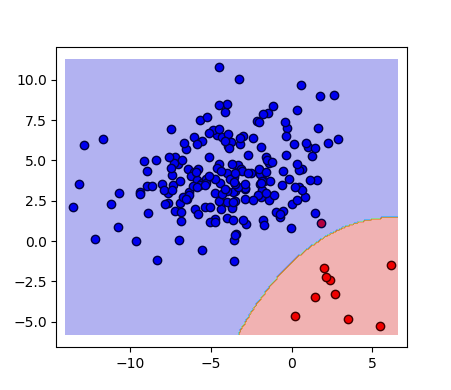}}%
\subfigure[][]{%
\includegraphics[height=2in]{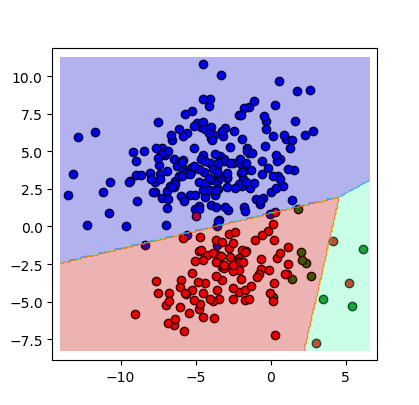}}%
\subfigure[][]{%
\includegraphics[height=2in]{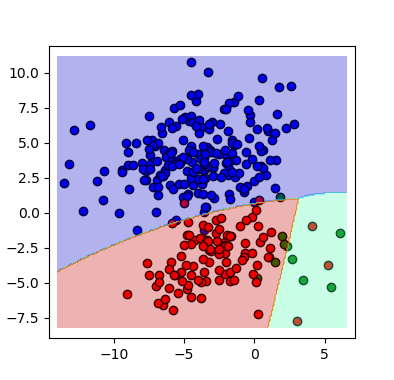}} \\
\subfigure[][]{%
\includegraphics[height=2in]{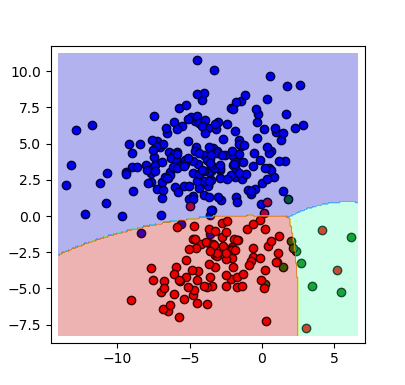}}
\subfigure[][]{%
\includegraphics[height=2in]{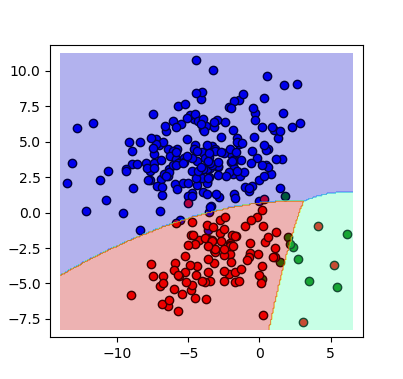}}%
\caption{Experiments with different class sample sizes: (a) LDA for two classes, (b) QDA for two classes, (c) Gaussian naive Bayes for two classes, (d) Bayes for two classes, (e) LDA for three classes, (f) QDA for three classes, (g) Gaussian naive Bayes for three classes, and (h) Bayes for three classes.}%
\label{figure_different_sizes}%
\end{figure*}

\begin{figure*}[!t]
% https://stackoverflow.com/questions/41041534/side-by-side-subfigure-in-sharelatex
\centering
\subfigure[][]{%
\includegraphics[height=2.1in]{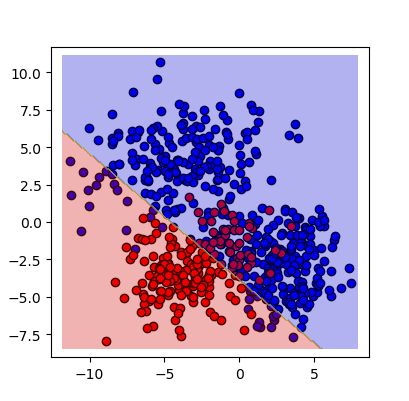}} 
\subfigure[][]{%
\includegraphics[height=2.1in]{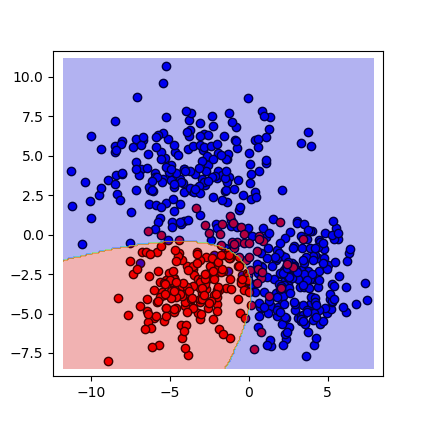}} \\
\hspace{0pt}%
\subfigure[][]{%
\includegraphics[height=2.1in]{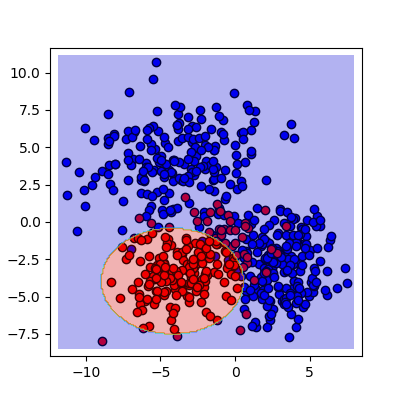}} 
\subfigure[][]{%
\includegraphics[height=2.1in]{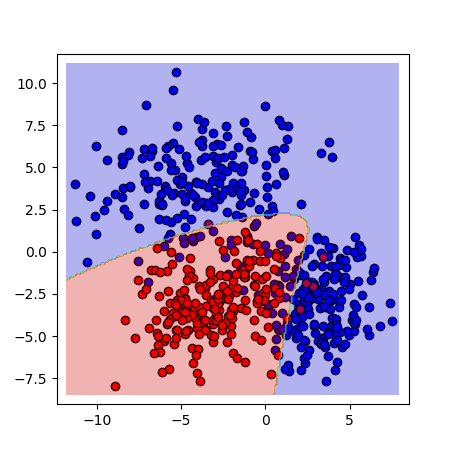}}%
\caption{Experiments with multi-modal data: (a) LDA, (b) QDA, (c) Gaussian naive Bayes, and (d) Bayes.}%
\label{figure_multi_modal}%
\end{figure*}

\section{Simulations}

In this section, we report some simulations which make the concepts of tutorial clearer by illustration.

\subsection{Experiments with Equal Class Sample Sizes}

We created a synthetic dataset of three classes each of which is a two dimensional Gaussian distribution. The means and covariance matrices of the three Gaussians from which the class samples were randomly drawn are:
\begin{align*}
& \b{\mu}_1 = [-4,4]^\top, ~~ \b{\mu}_2 = [3,-3]^\top, ~~ \b{\mu}_1 = [-3,3]^\top, \\
& \b{\Sigma}_1 = 
\begin{bmatrix}
    10 & 1 \\
    1 & 5 
\end{bmatrix}, ~~
\b{\Sigma}_2 = 
\begin{bmatrix}
    3 & 0 \\
    0 & 4 
\end{bmatrix}, ~~
\b{\Sigma}_3 = 
\begin{bmatrix}
    6 & 1.5 \\
    1.5 & 4 
\end{bmatrix}.
\end{align*}
The three classes are shown in Fig. \ref{figure_datasets}-a where each has sample size $200$.
Experiments were performed on the three classes. We also performed experiments on two of the three classes to test a binary classification. The two classes are shown in Fig. \ref{figure_datasets}-b.  
The LDA, QDA, naive Bayes, and Bayes classifications of the two and three classes are shown in Fig. \ref{figure_equal_sizes}.
For both binary and ternary classification with LDA and QDA, we used Eqs. (\ref{LDA_multiple_class_delta}) and (\ref{QDA_multiple_class_delta}), respectively, with Eq. (\ref{equation_max_delta}).
We also estimated the mean and covariance using Eqs. (\ref{equation_LDA_mean_estimate}), (\ref{equation_LDA_covariance_estimate}), and (\ref{equation_LDA_covariance_estimate_unified}).
For Gaussian naive Bayes, we used Eqs. (\ref{equation_naive_bayes_classification}) and (\ref{equation_Gaussian_naive_bayes_prob}) and estimated the parameters using Eqs. (\ref{equation_naiveBayes_mean_estimate}) and (\ref{equation_naiveBayes_covariance_estimate}).
For Bayes classifier, we used Eq. (\ref{equation_bayes_classification_2}) with Eq. (\ref{equation_Gaussian_naive_bayes_prob}) but we do not estimate the mean and variance; except, in order to use the exact likelihoods in Eq. (\ref{equation_bayes_classification_2}), we use the exact mean and covariance matrices of the distributions which we sampled from. We, however, estimated the priors.
The priors were estimated using Eq. (\ref{equation_prior_Bernoulli}) for all the classifiers.

As can be seen in Fig. \ref{figure_equal_sizes}, the space is partitioned into two/three parts and this validates the assertion that LDA and QDA can be considered as metric learning methods as discussed in Section \ref{section_metric_learning}.
As expected, the boundaries of LDA and QDA are linear and curvy (quadratic), respectively. 
The results of QDA, Gaussian naive Bayes, and Bayes are very similar although they have slight differences. This is because the classes are already Gaussian so if the estimates of means and covariance matrices are accurate enough, QDA and Bayes are equivalent. The classes are Gaussians and the off-diagonal elements of covariance matrices are also small compared to the diagonal; therefore, naive Bayes is also behaving similarly.

\subsection{Experiments with Small Class Sample Sizes}

According to Monte-Carlo approximation \cite{robert2013monte}, the estimates in Eqs. (\ref{equation_LDA_mean_estimate}), (\ref{equation_LDA_covariance_estimate}), (\ref{equation_naiveBayes_mean_estimate}) and (\ref{equation_naiveBayes_covariance_estimate}) are more accurate if the sample size goes to infinity, i.e., $n \rightarrow \infty$.
Therefore, if the sample size is small, we expect mode difference between QDA and Bayes classifiers. 
We made a synthetic dataset with three or two classes with the same mentioned means and covariance matrices. The sample size of every class was $10$.
Figures \ref{figure_datasets}-c and \ref{figure_datasets}-d show these datasets.
The results of LDA, QDA, Gaussian naive Bayes, and Bayes classifiers for this dataset are shown in Fig. \ref{figure_small_sizes}.
As can be seen, now, the results of QDA, Gaussian naive Bayes, and Bayes are different for the reason explained.

\subsection{Experiments with Different Class Sample Sizes}

According to Eq. (\ref{equation_prior_Bernoulli}) used in Eqs. (\ref{QDA_multiple_class_delta}), (\ref{LDA_multiple_class_delta}), (\ref{equation_bayes_classification_2}), and (\ref{equation_naive_bayes_classification}), the prior of a class changes by the sample size of the class. 
In order to see the effect of sample size, we made a synthetic dataset with different class sizes, i.e., $200$, $100$, and $10$, shown in Figs. \ref{figure_datasets}-e, \ref{figure_datasets}-f. We used the same mentioned means and covariance matrices.
The results are shown in Fig. \ref{figure_different_sizes}.
As can be seen, the class with small sample size has covered a small portion of space in discrimination which is expected because its prior is small according to Eq. (\ref{equation_prior_Bernoulli}); therefore, its posterior is small. On the other hand, the class with large sample size has covered a larger portion because of a larger prior.

\subsection{Experiments with Multi-Modal Data}

As mentioned in Section \ref{section_FDA}, LDA and QDA assume uni-modal Gaussian distribution for every class and thus FDA or LDA faces problem for multi-modal data \cite{sugiyama2007dimensionality}.
For testing this, we made a synthetic dataset with two classes, one with sample size $400$ having two modes of Gaussians and the other with sample size $200$ having one mode. We again used the same mentioned means and covariance matrices.
The dataset is shown in Fig. \ref{figure_datasets}-g.

The results of the LDA, QDA, Gaussian naive Bayes, and Bayes classifiers for this dataset are shown in Fig. \ref{figure_multi_modal}. 
The mean and covariance matrix of the larger class, although it has two modes, were estimated using Eqs. (\ref{equation_LDA_mean_estimate}), (\ref{equation_LDA_covariance_estimate}), (\ref{equation_naiveBayes_mean_estimate}) and (\ref{equation_naiveBayes_covariance_estimate}) in LDA, QDA, and Gaussian naive Bayes.
However, for the likelihood used in Bayes classifier, i.e., in Eq. (\ref{equation_bayes_classification_2}), we need to know the exact multi-modal distribution. Therefore, we fit a mixture of two Gaussians \cite{ghojogh2019fitting} to the data of the larger class:
\begin{align}
\mathbb{P}(X=\b{x} \,|\, \b{x} \in \mathcal{C}_k) = \sum_{k=1}^2 w_k\, f(\b{x}; \b{\mu}_k, \b{\Sigma}_k),
\end{align}
where $f(\b{x}; \b{\mu}_k, \b{\Sigma}_k)$ is Eq. (\ref{equation_multivariate_Gaussian}) and we the fitted parameters were:
\begin{align*}
&\b{\mu}_1 = [-3.88, 4]^\top, ~~ \b{\mu}_2 = [3.04, -2.92]^\top, \\
&\b{\Sigma}_1 = 
\begin{bmatrix}
    9.27 & 0.79 \\
    0.79 & 4.82 
\end{bmatrix}, ~~
\b{\Sigma}_2 = 
\begin{bmatrix}
    2.87 & 0.03 \\
    0.03 & 3.78 
\end{bmatrix}, \\
& w_1 = 0.49, ~~ w_2 = 0.502.
\end{align*}
As Fig. \ref{figure_multi_modal} shows, LDA has not performed well enough as expected. The performance of QDA is more acceptable than LDA but still not good enough because QDA also assumes a uni-modal Gaussian for every class. The result of Gaussian naive Bayes is very different from the Bayes here because the Gaussian naive Bayes assumes uni-modal Gaussian with diagonal covariance for every class.
Finally, the Bayes has the best result as it takes into account the multi-modality of the data and it is optimum \cite{mitchell1997machine}.

\section{Conclusion and Future Work}

This paper was a tutorial paper for LDA and QDA as two fundamental classification methods. We explained the relations of these two methods with some other methods in machine learning, manifold (subspace) learning, metric learning, statistics, and statistical testing. Some simulations were also provided for better clarification.

This paper focused on LDA and QDA which are discriminators with one and two polynomial degrees of freedom, respectively.
As the future work, we will work on a tutorial paper for non-linear discriminant analysis using kernels \cite{baudat2000generalized,li2003recognising,lu2003face}, which is called \textit{kernel discriminant analysis}, to have discriminators with more than two degrees of freedom.

\section*{Acknowledgment}
The authors hugely thank Prof. Ali Ghodsi (see his great online related courses \cite{web_classification,web_data_visualization}), Prof. Mu Zhu, Prof. Hoda Mohammadzade, and other professors whose courses have partly covered the materials mentioned in this tutorial paper.

% \clearpage
% \pagebreak

% \newpage
% \newpage

\bibliography{References}
\bibliographystyle{icml2016}

\end{document}